\documentclass{article}

\PassOptionsToPackage{numbers,compress}{natbib}
\usepackage[preprint]{neurips_2026}

\usepackage[utf8]{inputenc}
\usepackage[T1]{fontenc}
\usepackage{hyperref}
\usepackage{url}
\usepackage{booktabs}
\usepackage{amsfonts}
\usepackage{nicefrac}
\usepackage{microtype}
\usepackage{xcolor}
\usepackage{graphicx}
\usepackage{tabularx}
\usepackage{makecell}
\usepackage{bm}
\usepackage{amsmath}
\usepackage{amssymb}
\usepackage{amsthm}
\usepackage{multirow}
\usepackage{array}

\newtheorem{manualtheoreminner}{Principle}
\newenvironment{manualtheorem}[1]{%
  \renewcommand\themanualtheoreminner{#1}%
  \manualtheoreminner
}{\endmanualtheoreminner}

\title{A Reconstruction-Based Framework for Caption Evaluation Beyond Reference Captions}

\author{%
  Zhijiang Tang$^{2,*}$ \enspace
  Jiaxin Qi$^{1,*}$ \enspace
  Kaihua Tang$^{3}$ \enspace
  Yuhua Zheng$^{2}$ \enspace
  Jianqiang Huang$^{1,2,\dagger}$\\
  $^{1}$Computer Network Information Center, Chinese Academy of Sciences, China\\
  $^{2}$Hangzhou Institute for Advanced Study, University of Chinese Academy of Sciences, China\\
  $^{3}$Tongji University, China\\
  \texttt{tangzhijiang24@mails.ucas.ac.cn, jxqi@cnic.cn, tangkaihua@tongji.edu.cn}\\
  \texttt{zhengyuhua@ucas.ac.cn, jqhuang@cnic.cn}\\
  $^{*}$Equal contribution \quad $^{\dagger}$Corresponding author
}

\hypersetup{
  hidelinks,
  pdftitle={A Reconstruction-Based Framework for Caption Evaluation Beyond Reference Captions},
  pdfauthor={Zhijiang Tang, Jiaxin Qi, Kaihua Tang, Yuhua Zheng, Jianqiang Huang}
}

\begin{document}

\maketitle
\begin{abstract}

Image captioning is a primary task in vision--language research, yet assessing how faithfully a caption preserves image semantics without relying on reference captions remains unsettled. 
Prevailing evaluations rely on human-annotated references, whose content reflects annotator intent and captioning proficiency. 
In this paper, we study a reconstruction-based principle for caption evaluation: a caption is as good as its capacity to enable reconstruction of the original image.
However, because captioning inherently compresses visual information, it is impossible to recover all details, and pixel-wise comparison between reconstructed and source images is neither feasible nor meaningful. 
Through our in-depth analysis of the nature of captions, whose fundamental purpose is to transmit the semantic content of an image, we propose a revised principle: a caption is as good as its capacity to enable a reconstruction that is semantically equivalent to the original.
To assess semantic equivalence, we test whether the reconstruction matches the original image across a suite of downstream vision--language tasks, yielding a reference-free, task-conditioned caption score. 
We characterize component-dependent limitations and introduce the lower-cost Captioning Turing Test Dataset (CTTD) surrogate. 

\end{abstract}


\section{Introduction}
\label{sec:intro}

\begin{figure}[t!]
\centering\includegraphics[width=\linewidth]{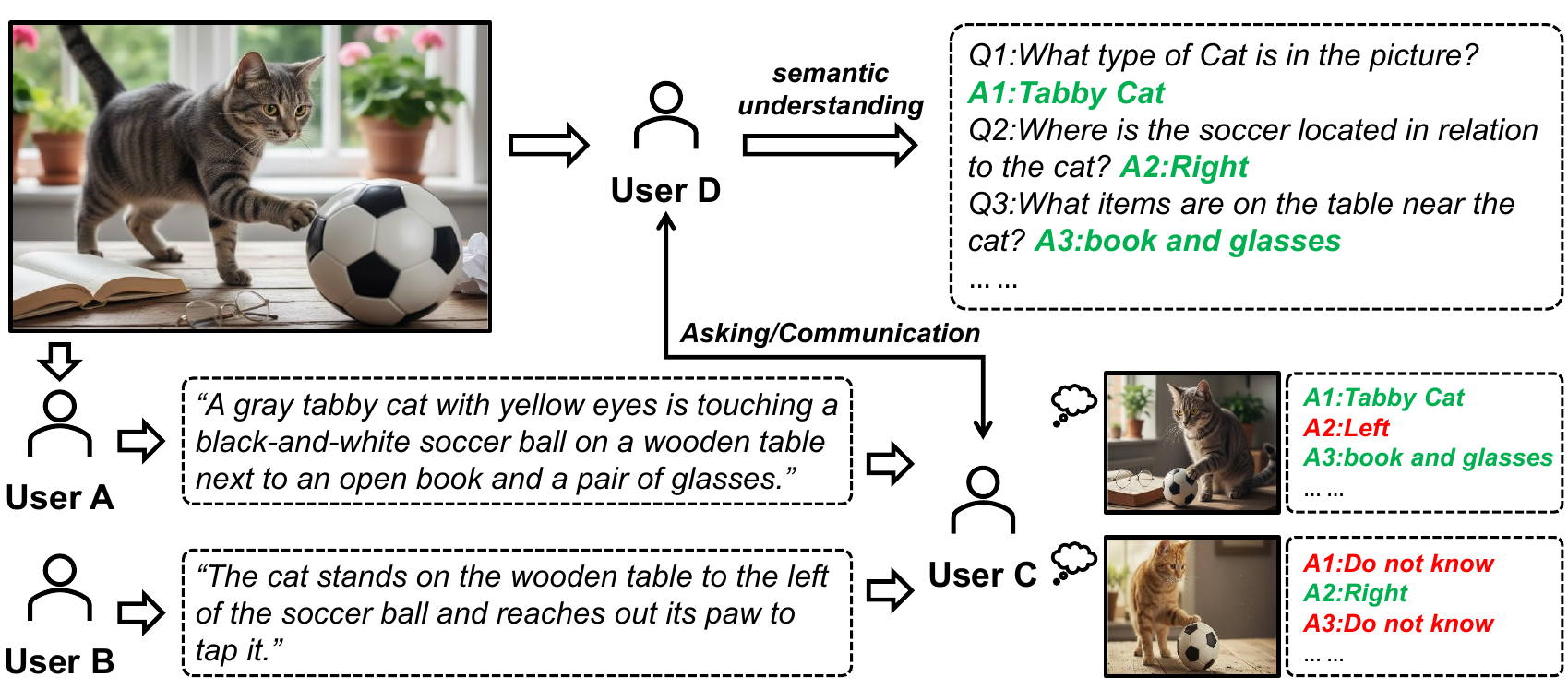}
    \caption{Illustration of the fundamental role of captions, conveying semantics of an image to others. Users A and B, with different annotation preferences, produce captions that emphasize different facts, leading User C to imagine different scenes. A good caption minimizes the semantic gap across image viewers and caption readers, which is illustrated as question--answer agreement between image viewer (User D) and caption recipient (User C).}
    \label{fig:teaser}
\end{figure}

Image captioning generates natural language descriptions of visual content and is fundamental to vision--language research~\cite{herdade2019image,hossain2019comprehensive}. Despite progress in caption models~\cite{li2023blip}, the community still needs reference-free signals for how faithfully a caption preserves image semantics~\cite{berger2024surveying}. Current protocols largely compare generated captions with human-written references~\cite{lai2024revisit,sarto2025image}, whose coverage reflects annotator intent and captioning proficiency. 
For example, to caption the image in Figure~\ref{fig:teaser}, some annotators may prefer
long and fine-grained captions with attributes and contextual details (e.g., ``A gray tabby cat with yellow eyes is touching a black-and-white soccer ball on a wooden table next to an open book and a pair of glasses''); others may emphasize spatial relations and interactions (e.g., ``The cat stands on the wooden table to the left of the soccer ball and reaches out its paw to tap it''). Such differences make captioning task-dependent: human preferences remain important, while a separate reference-free signal can measure semantic preservation.

To develop such a signal, we revisit the fundamental role of the caption. Consider user A writing a caption to convey an image to user C: if C can mentally form the same scene that A saw, the caption has succeeded. This suggests a reconstruction principle: \textit{a caption is as good as its capacity to enable reconstruction of the original image}. Literal enforcement is impractical because a caption compresses visual information and cannot transmit every detail. Moreover, even a semantically faithful reconstruction can shift an object and incur a large pixel-level error. Exact visual reconstruction therefore cannot be the evaluation strategy.

To make this principle operational, we further consider the practical role of the caption: bridging the visual information gap between users for downstream understandings. 
The caption does not need to reproduce the image at the pixel level; it is successful as long as different users can imagine the scene without the semantic mismatch.
For example, even if user C does not imagine the exact source image based on user A's caption in Figure~\ref{fig:teaser}, C can still engage in the following interaction and share the same understanding with user D about the image.
This motivates a refined principle for caption evaluation: \textit{a caption is as good as its capacity to enable a reconstruction that is semantically equivalent to the original}. Because semantic equivalence is not directly observable, we approximate it by comparing performance on downstream tasks such as visual question answering, grounding, and reasoning. The resulting score is a task-conditioned diagnostic of preserved semantics rather than a universal definition of caption quality.

We conduct experiments and ablations to characterize the framework, including its sensitivity to text-to-image generators. Because exhaustive testing is costly and each task probes only part of a caption, we construct the Captioning Turing Test Dataset (CTTD), which assigns a compact set of categorized questions to curated images. CTTD reproduces broad trends of the full task suite at substantially lower cost, while both evaluations remain dependent on the generator, judge, and question selection. We use original-image performance as an internal upper reference, not as a substitute for direct validation against human caption judgments.
By measuring semantic retention across vision-language tasks, our framework supplies a reference-free signal that complements human evaluation and preference-oriented reference metrics.

Our contribution can be summarized in three aspects:
\begin{itemize}
\item We revisit current caption evaluation practices and propose a reconstruction-based principle that complements reference-based metrics, and further make this principle operational by introducing the semantic equivalence between the reconstructed image and the source image.
\item Building on this principle, we propose a reference-free protocol that evaluates whether a caption-conditioned reconstruction supports the original image's downstream tasks. We then use experiments and ablations to characterize the implementation and its limitations.
\item We release the Captioning Turing Test Dataset, a scalable and comprehensive benchmark that serves as a practical surrogate for the exhaustive testing above, enabling efficient and reference-free caption assessment.
\end{itemize}

\textbf{Code: }\href{https://github.com/ZhijiangTang/Caption-Turing-Test}{https://github.com/ZhijiangTang/Caption-Turing-Test}

\section{Related Work}

\begin{figure*}[t]
    \centering
    \includegraphics[width=\textwidth]{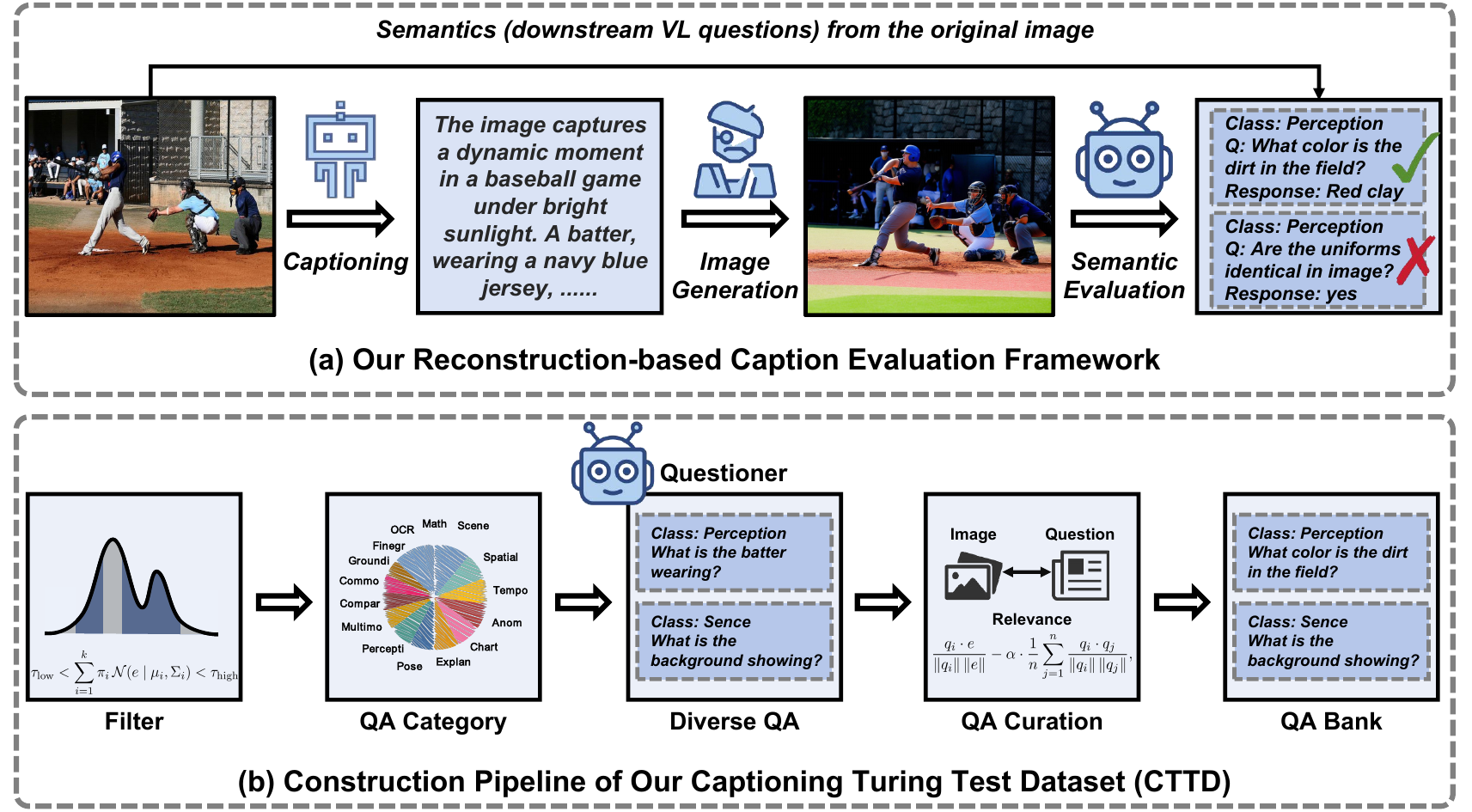}
    \caption{Illustration of our framework and the Captioning Turing Test Dataset (CTTD) pipeline. 
    \textbf{(a)} Overview of our framework: The generator reconstructs the image based on the generated caption, and the judger evaluates the reconstructed image by answering related questions. Response accuracy is used as a score to assess the quality of the caption.
    \textbf{(b)} Construction pipeline of our CTTD. After image filtering, we aggregate vision-language tasks into question and answer (QA) categories. The questioner generates category-conditioned QAs, and curation produces the final QA bank.
    }
    \label{fig:method}
\end{figure*}

\subsection{Image Captioning Model}
Image captioning has progressed through several major architectures. ShowAndTell~\cite{vinyals2015show} combined convolutional and recurrent neural networks to bridge vision and language, while PureT~\cite{wang2022end} directly generated captions from image features using a pure Transformer~\cite{vaswani2017attention}.

Pre-trained VLMs~\cite{tang2026cccaption,radford2021learning,li2022blip,li2023blip} provided new approaches to caption generation. The success of LLMs~\cite{brown2020language} further enabled LVLMs such as the Qwen~\cite{qwen2_5_vl,qwen3_vl} and LLaVa~\cite{liu2023visual} series, whose perception and reasoning abilities improve caption semantics. This progress motivates distinct semantic-preservation signals.

\subsection{Image Caption Evaluation}

\noindent\textbf{LVLM-based}. FLEUR~\cite{fleur} directly scores captions with an LVLM, CapArena~\cite{caparena} compares caption pairs, and PROMETHEUS-VISION~\cite{Prometheus-vision} fine-tunes an LVLM evaluator. CapRL~\cite{caprl} instead asks an LLM to answer vision-related questions from a caption and uses answer accuracy as the score. Although LVLMs bridge images and captions, their language bias can still distort evaluation results.

\noindent\textbf{Semantic-based}.
Semantic metrics compare cross-modal representations such as CLIP~\cite{clip} or probe generated content. CLIPScore~\cite{clipscore} and PAC-S~\cite{sarto2023pacs} score image--caption embedding alignment, whereas CAMScore~\cite{CAMScore} and Image2Text2Image~\cite{image2text2image} use caption-conditioned image reconstruction. VisualFactChecker~\cite{VisualFactChecker} is a detailed-caption generation and fact-checking pipeline that separately introduces a CLIP-based source--reconstruction score, rather than a pixel-level comparison. TIFA~\cite{hu2023tifa} and Davidsonian Scene Graph evaluation~\cite{cho2024dsg} use question answering to assess text-to-image faithfulness. Our downstream-task score differs operationally but retains generator--judge dependence.

\section{Method}
\label{sec:method}

\subsection{Preliminaries}

Formally, we denote the space of images by $\mathcal{X}$ and the space of captions by $\mathcal{C}$.
An image captioning model is a mapping $f_{\text{cap}} : \mathcal{X} \!\rightarrow\! \mathcal{C}$, which maps an input image $I \!\in\! \mathcal{X}$ to a caption $C = f_{\text{cap}}(I)$. In practice, a caption is represented as a finite token sequence $C = (c_1,\dots,c_T)$ with length $T$. The mapping $f_{\text{cap}}$ is implemented by a parametric model with parameters $\theta$ that defines a
conditional distribution $p_{\theta}(C \!\mid\! I)$.
The model is trained on image--caption pairs by minimizing the negative
log-likelihood (NLL):
\begin{equation}
\label{loss_nll}
    \mathcal{L}_{\text{NLL}}(I, C; \theta)
    = - \sum_{t=1}^{T} \log p_{\theta}\big(c_t \mid c_{<t}, I\big),
\end{equation}
where $c_{<t} = (c_1,\dots,c_{t-1})$ denotes the caption prefix.
This objective trains $f_{\text{cap}}$ to imitate reference captions, but it does not assess how well a caption preserves the image’s visual information. For example, COCO captions~\cite{lin2014microsoft} are typically short, so longer generated captions are often penalized when compared to them, even though they may convey richer and higher-quality descriptions.

To complement reference-based metrics without requiring human-written captions, we propose the following reconstruction-based principle for caption evaluation.

\begin{manualtheorem}{1} For a fixed image $I$, among captions $C \in \mathcal{C}$, a caption $C_1$ is considered better than $C_2$ if the image reconstructed from $C_1$ is more similar to $I$ than the image reconstructed from $C_2$.
\end{manualtheorem}

In practice, the principle can be instantiated by defining a caption-to-image mapping $f_{\text{img}} : \mathcal{C} \rightarrow \mathcal{X}$ based on a frozen text-to-image generator. Then, we obtain a reconstruction $\hat{I} = f_{img}(C)$ and assign a caption evaluation score:
\begin{equation}
    \label{eq:score}
    S(I,C) = \Phi\big(I, \hat{I}=f_{\text{img}}(C)\big),
\end{equation}
where $\Phi : \mathcal{X} \times \mathcal{X} \rightarrow \mathbb{R}$ is an image similarity function and $S(\cdot)$ is the scoring function. The most intuitive choice for $\Phi$ would be the pixel-level $\ell_2$ distance between $I$ and $\hat{I}$. However, this is impractical for the reasons discussed above. Consequently, a practical reference-free scoring rule remains challenging.

\subsection{Our Framework}

Based on the above discussion, a caption can be viewed as a medium for transmitting the semantic content of an image rather than its exact pixels. To formalize this, we propose to use a semantic equivalence relation on the image space $\mathcal{X}$, where $I_1 \sim I_2$ indicates that the two images are semantically interchangeable for downstream understanding. For a given image $I \in \mathcal{X}$, its semantic equivalence class is:
\begin{equation}
\label{eq:semantic_eq}
    [I] = \{ I' \in \mathcal{X} \mid I' \sim I \},
\end{equation}
which collects all images semantically equivalent to $I$. Under this formulation, the reconstruction target is relaxed from reproducing $I$ exactly to generating any image $\hat{I} \in [I]$ in the same semantic equivalence class. This leads to the following refined reconstruction-based evaluation principle.

\begin{manualtheorem}{2} For a fixed image $I$, consider captions $C \in \mathcal{C}$, a caption $C_1$ is considered better than $C_2$ if the reconstructed image $\hat{I}_1$ lies closer to the semantic equivalence class $[I]$ than $\hat{I}_2$ does.
\end{manualtheorem}

However, the semantic equivalence defined above is not directly observable. As shown in Figure~\ref{fig:method}(a), we instead approximate it with a suite of downstream vision--language evaluations: if replacing the original image with a reconstructed image allows a third-party judge to achieve comparable performance across these benchmarks, the two images are treated as semantically equivalent. Let $\mathcal{D} \!= \! \{ \mathcal{D}_i \}_{i=1}^{n}$ denote a collection of vision--language datasets, such as visual question answering~\cite{vqa} and visual grounding~\cite{zhang2025beyond}. A fixed judge model $\mathcal{J}$ is used to perform each task, and the semantic equivalence between $I$ and $\hat{I}$ under the task suite $\mathcal{D}$ is defined as:
\begin{equation}
\label{eq:vl_equiv}
    I \sim_{\sigma} \hat{I}
    \;\text{iff}\;
    \frac{1}{n} \sum_{i=1}^{n}
    \bigl| \mathcal{J}(I, \mathcal{D}_i) - \mathcal{J}(\hat{I}, \mathcal{D}_i) \bigr|
    \le \sigma,
\end{equation}
where $\mathcal{J}(I, \mathcal{D}_i)$ is the performance (e.g., accuracy) of the judge model on dataset $\mathcal{D}_i$ when conditioned on image $I$, $\sigma > 0$ is a tolerance, and a smaller $\sigma$ induces a stricter notion of semantic equivalence.

In practice, original-image performance is an upper reference at the population level on CTTD, motivating the use of judge scores on $\hat{I}$ as a ranking surrogate. Equation~\eqref{eq:vl_equiv} states the intended equivalence criterion, whereas Eq.~\eqref{eq:vl_score} is a practical score and is not algebraically equivalent to that criterion. A simplified reconstruction may be easier for the judge, and generator stochasticity may alter layout; the score can therefore conflate caption fidelity with generator and judge behavior. Under fixed components, a caption receives a higher score when its reconstruction attains higher downstream-task performance:
\begin{equation}
\label{eq:vl_score}
    S_{\mathcal{D}}(I,C)
    = \frac{1}{n} \sum_{i=1}^{n}
      \mathcal{J}\big(\hat{I}=f_{\text{img}}(C), \mathcal{D}_i\big),
\end{equation}
where original-image terms are constant for comparisons on the same image, although omitting them does not remove the component confounds above.
However, Eq.~\eqref{eq:vl_score} is computationally expensive, as it requires running
a high-capacity text-to-image generator $f_{\text{img}}$ and a strong vision--language model $\mathcal{J}$ over multiple benchmarks for every candidate caption. Moreover, each $\mathcal{D}_i$ only probes a narrow aspect of semantic equivalence (e.g., visual grounding mainly tests localization). We therefore construct a compact surrogate.

\subsection{Captioning Turing Test Dataset (CTTD)}

\begin{table*}[t]
  \centering
    \renewcommand{\arraystretch}{1.4}
    \setlength{\tabcolsep}{1.5pt}
\begin{tabularx}{\textwidth}{l|>{\centering\arraybackslash}X>{\centering\arraybackslash}X>{\centering\arraybackslash}X>{\centering\arraybackslash}X>{\centering\arraybackslash}X>{\centering\arraybackslash}X>{\centering\arraybackslash}X>{\centering\arraybackslash}X>{\centering\arraybackslash}X>{\centering\arraybackslash}X>{\centering\arraybackslash}X>{\centering\arraybackslash}X>{\centering\arraybackslash}X}
    \toprule
    \multicolumn{1}{c|}{\multirow{2}{*}{{Captioning Model}}} & MMB.  & MMS.  & Hall. & MME   & Vista. & Verse. & Vision. & OB. & OQ. & CO. &\multicolumn{1}{c}{\multirow{2}{*}{\textbf{Avg.}}}  \\
    & \citeyear{mmbench}  &  \citeyear{mmstar}  &\citeyear{hallusionbench} & \citeyear{mme}   & \citeyear{mathvista} & \citeyear{mathverse}  & \citeyear{MathVision} & \citeyear{ocrbench} & \citeyear{ocrvqa} & \citeyear{refcoco} & \\
    \midrule
    Upperbound  & 83.97  & 46.73  & 48.48  & 83.91  & 51.20  & 28.53  & 20.50  & 68.50  & 54.38  & 36.48  & 52.27  \\
    \midrule
    ShowAndTell~\cite{vinyals2015show} & 41.79  & 21.00  & 2.31  & 53.12  & 29.60  & 12.08  & 7.79  & 0.90  & 20.02  & 19.17  & 20.78  \\
    BLIP~\cite{li2022blip}  & 66.92  & 32.60  & 17.46  & 64.36  & 34.30  & 11.88  & 7.68  & 7.00  & 20.69  & 19.07  & 28.20  \\
    BLIP2~\cite{li2023blip} & 68.77  & 33.33  & 18.61  & 65.21  & 34.80  & 11.78  & 7.89  & 9.10  & 17.42  & 19.26  & 28.62  \\
    LLaVa-V1.6-7B~\cite{liu2024improved} & 74.73  & 38.13  & 34.49  & 73.97  & 38.60  & 22.23  & 17.65  & 23.83  & 44.85  & 19.80  & 38.83  \\
    LLaVa-V1.6-34B~\cite{liu2024improved} & 76.21  & 38.33  & 37.33  & 74.60  & 39.00  & 22.74  & \textbf{19.63 } & 21.03  & 48.12  & 20.48  & 39.75  \\
    Llama-3.2-11B~\cite{llama3_2_11b_vision} & 78.47  & 41.20  & 39.85  & \underline{76.92}  & 41.60  & 23.05  & 18.97  & 26.92  & 45.91  & 21.62  & 41.45  \\
    Qwen2.5-VL-3B~\cite{qwen2_5_vl} & 75.21  & 39.20  & 38.59  & 75.44  & 41.70  & 22.94  & 18.53  & 26.61  & 43.98  & 22.09  & 40.43  \\
    Qwen2.5-VL-7B~\cite{qwen2_5_vl} & 78.66  & 41.27  & \underline{43.22}  & 76.20  & 42.60  & 23.65  & 18.75  & 30.53  & 49.09  & 22.58  & 42.65  \\
    Qwen2.5-VL-72B~\cite{qwen2_5_vl} & 79.63  & 42.00  & 42.06  & 76.87  & 42.40  & 24.06  & \underline{19.52 } & 29.79  & \textbf{50.53 } & 22.84  & 42.97  \\
    InternVL3.5-8b~\cite{internvl3_5} & 78.96  & 42.53  & 41.96  & 76.16  & \underline{44.10}  & 25.79  & 18.97  & 29.84  & 46.29  & 22.30  & 42.69  \\
    InternVL3.5-38b~\cite{internvl3_5} & 80.13  & 40.73  & \textbf{44.37 } & 76.71  & \textbf{44.80 } & \underline{26.29 } & 19.30  &\underline{ 33.43 } & 46.78  & 22.36  & \underline{43.49}  \\
    Qwen3-VL-2B~\cite{qwen3_vl} & \underline{80.46}  & \underline{42.60}  & 41.96  & \textbf{78.10 } & 43.30  & 23.96  & 17.98  & \textbf{35.57 } & 47.35  & \underline{23.18 } & 43.45  \\
    Qwen3-VL-32B~\cite{qwen3_vl} & \textbf{80.71 } & \textbf{43.47 } & 41.85  & \textbf{78.10 } & 43.20  & \textbf{27.21 } & 19.08  & 30.61  &\underline{ 49.57}  & \textbf{23.65 } & \textbf{43.74 } \\
    \bottomrule
\end{tabularx}%
  \caption{Performance of vision-language tasks across different captioning models on various datasets. Bold numbers indicate the best performance, and underlined numbers indicate the runner-up. ``Upperbound'' refers to the performance achieved by completing the task using the original image. ``Avg.'' represents the average performance across all datasets. We have provided detailed information about each dataset in the Section~\ref{sec:vlt}.
}
  \label{tab:vqa}%
\end{table*}

To make the above principle practical, we construct a compact QA-style benchmark with diverse questions per image, the \emph{Captioning Turing Test Dataset} (CTTD), for lower-cost reference-free evaluation. The name reflects an operational test: a caption passes when its reconstruction can substitute for the original for the selected downstream models. Human agreement requires separate validation.

\noindent\textbf{Image Selection.}
We collect images from diverse open-source datasets and remove duplicates and anomalies.
Following~\cite{zhang2025beyond}, we fit a Gaussian mixture model (GMM)~\cite{mclachlan2019finite} to image embeddings and retain images whose density falls within a target range:
\begin{equation}
    \tau_{\text{low}} 
    < \sum_{i=1}^{k} \pi_i \, \mathcal{N}(e \mid \mu_i, \Sigma_i) 
    < \tau_{\text{high}},
\end{equation}
where $e$ is the image embedding and $\{\pi_i, \mu_i, \Sigma_i\}_{i=1}^{k}$ are GMM parameters, $\tau_{\text{low}}$ and $\tau_{\text{high}}$ are threshold values.

\noindent\textbf{Question-Answer Generation.}
To cover diverse vision--language capabilities, we first build a pipeline for deriving question categories: For each image, an LVLM first proposes candidate questions. We then embed all questions and cluster them using DBSCAN~\cite{dbscan}, and ask the LVLM to summarize each cluster into a question category. 

To curate questions, for each image and question type, we prompt the LVLM to generate $m$ candidate QA pairs. We then (i) discard pairs with missing answers or answers flagged as inconsistent with the image during curation, (ii) remove near-duplicates with identical answers, and (iii) rank the remainder by a relevance--diversity score:
\begin{equation}
    s(q_i, e)
    = \frac{q_i \cdot e}{\|q_i\| \, \|e\|}
      - \alpha \cdot \frac{1}{n} \sum_{j=1}^{n}
        \frac{q_i \cdot q_j}{\|q_i\| \, \|q_j\|},
\end{equation}
where $q_i$ and $e$ are the embeddings of the $i$-th QA and the image, $n$ is the number of candidate QAs, and $\alpha$ balances relevance and redundancy. We retain the top-scoring QAs for each image, resulting in about 7k images with 15 questions per image on average; this automated filtering may favor questions that the curation components find easier.
Figure~\ref{fig:method}(b) summarizes the pipeline.
Section~\ref{exp} and the supplement provide details.

\section{Experiments}
\label{exp}

\begin{figure*}[!htbp]
    \centering
    \includegraphics[width=0.90\textwidth]{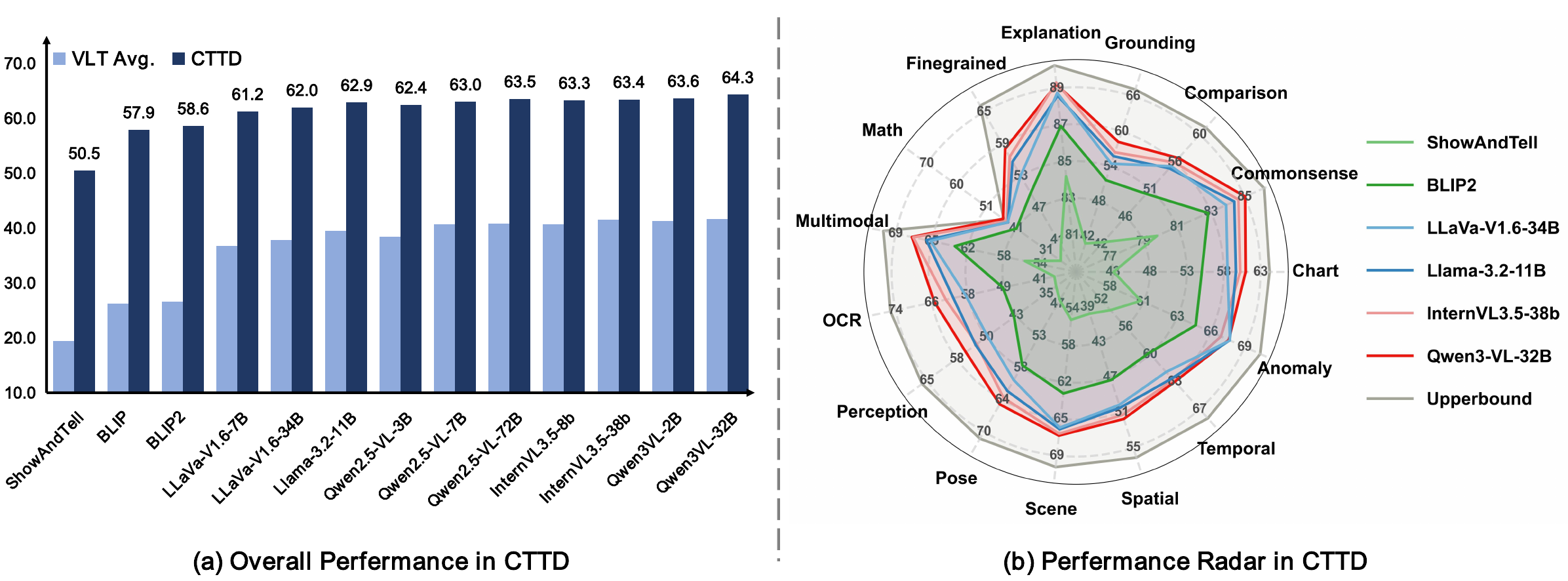}
    \caption{Performance of various captioning models in our Captioning Turing Test Dataset~(CTTD). \textbf{(a)} Overall performance, where ``VLT Avg.'' represents the average performance over vision-language tasks in Table~\ref{tab:vqa}. \textbf{(b)} Radar performance for different models across various question categories in CTTD, with a comparison to the upper bound.
}
    \label{fig:CTTD_overall}
\end{figure*}

\subsection{Vision Language Tasks}
\label{sec:vlt}

We categorize vision-language tasks into four classes. The following section introduces these tasks and their associated datasets.

\noindent\textbf{General Visual Question Answering~(VQA)} requires models to answer visual questions from an image, testing object recognition, scene interpretation, and reasoning. MMBench~(MMB.)~\cite{mmbench} provides bilingual multiple-choice questions covering a broad range of fine-grained evaluation dimensions; MMStar~(MMS.)~\cite{mmstar} contains 1500 carefully curated vision-indispensable samples; HallusionBench~(Hall.)~\cite{hallusionbench} uses 951 samples to probe hallucination, visual consistency, and reasoning; and MME~\cite{mme} evaluates perceptual and cognitive abilities across 14 sub-tasks.

\noindent\textbf{Science, Technology, Engineering, and Mathematics~(STEM)} emphasizes reasoning and answering in visual contexts such as charts and geometric figures. MathVista~(Vista.)~\cite{mathvista} aggregates 6141 multimodal math problems from 28 public and newly created datasets. MathVerse~(Verse.)~\cite{mathverse} contains 2612 illustrated math problems emphasizing visual--linguistic integration, while MathVision~(Vision.)~\cite{MathVision} contains 3040 real-world competition problems spanning 16 mathematical disciplines and 5 difficulty levels.

\noindent\textbf{Optical Character Recognition~(OCR)} requires models to recognize text, answer questions, and extract key information. OCRBench~(OB.)~\cite{ocrbench} spans 29 sub-datasets covering text recognition and key-information extraction, while OCRVQA~(OQ.)~\cite{ocrvqa} contains roughly 207,572 images and requires models to answer questions by reading in-image text, with 5\% sampled for evaluation.

\noindent\textbf{Visual Grounding} focuses on locating objects in images from textual descriptions and therefore assesses spatial understanding. We use RefCOCO~(CO.)~\cite{refcoco}; because reconstruction can shift or mirror layouts, absolute-coordinate IoU is a generator-sensitive diagnostic rather than a pure measure of caption fidelity.
Additional dataset details and task definitions are provided in the supplementary material.

\subsection{Implementation Details}

\noindent\textbf{Evaluation Settting}.
We evaluated the traditional captioning model ShowAndTell~\cite{vinyals2015show}; VLMs including BLIP~\cite{li2022blip} and BLIP2~\cite{li2023blip}; and LVLMs including LLaVa-V1.6 7B and 34B~\cite{liu2024improved}, Llama-3.2-vision 11B~\cite{llama3_2_11b_vision}, InternVL3.5 8B and 38B~\cite{internvl3_5}, Qwen2.5-VL 3B, 7B, and 72B~\cite{qwen2_5_vl}, and Qwen3-VL 2B and 32B~\cite{qwen3_vl}.

We used Qwen-Image~\cite{wu2025qwenimagetechnicalreport} as the reconstruction model $f_{\text{img}}$. For grounding, we used Intersection over Union~(IoU)~\cite{yu2016unitbox} with Qwen2.5-VL 72B as judge $\mathcal{J}$; this score jointly reflects captioned spatial relations, generated layout, and judge localization. For all other tasks, we used accuracy with Qwen3-VL 8B as the judge. All experiments were conducted on NVIDIA H100 GPUs.

\begin{figure*}[t]
    \centering
    \includegraphics[width=\textwidth]{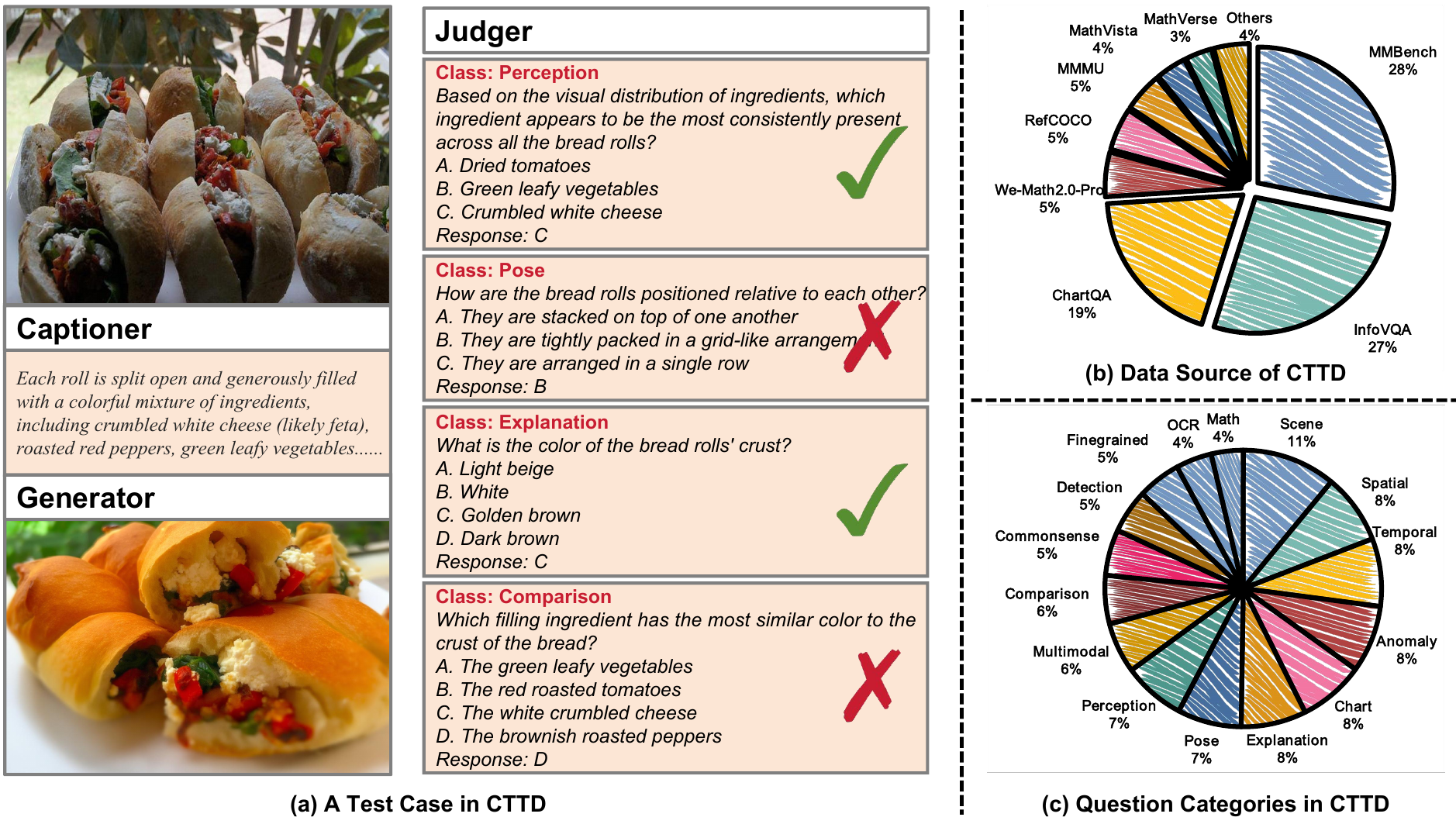}
    \caption{Overview of the Captioning Turing Test Dataset (CTTD). \textbf{(a)} A test case in CTTD. ``Captioner'', ``Generator'', and ``Judger'' correspond to Qwen2.5-VL-3B~\cite{qwen2_5_vl}, Qwen-Image~\cite{wu2025qwenimagetechnicalreport}, and Qwen3-VL-8B~\cite{qwen3_vl}, respectively. \textbf{(b)} The percentage of data sources used in CTTD. (c) The percentage of question categories in CTTD.
}
    \label{fig:statistical_analysis}
\end{figure*}
\noindent\textbf{Captioning Turing Test Dataset}.
We collected images from 9 publicly available datasets and filtered them through the GMM filter and human expert review, resulting in 7k images. Qwen2.5-VL 72B generated the initial questions, which were encoded using Ops-MM-embedding~\cite{ops_mm_embedding_v1_7B}. ChatGPT5~\cite{openai2025gpt5} then summarized clustered representative questions from each category, yielding 15 categories: perception, scene, spatial, temporal, comparison, commonsense, math, chart, OCR, detection, pose, fine-grained, anomaly, multimodal, and explanation.
Qwen3-VL 32B~\cite{qwen3_vl} generated 5 QAs for every image across each question category. During question curation, we reused Ops-MM-embedding~\cite{ops_mm_embedding_v1_7B} and set the relevance--diversity parameter to $\alpha=0.5$. The resulting CTTD contains 7k samples, with an average of 15 questions per sample. Further implementation details are provided in the supplementary material.

\begin{table*}[t]
  \centering
    \renewcommand{\arraystretch}{1.1}
\begin{tabularx}{\textwidth}{l|>{\centering\arraybackslash}X>{\centering\arraybackslash}X>{\centering\arraybackslash}X>{\centering\arraybackslash}X>{\centering\arraybackslash}X>{\centering\arraybackslash}X>{\centering\arraybackslash}X>{\centering\arraybackslash}X>{\centering\arraybackslash}X>{\centering\arraybackslash}X>{\centering\arraybackslash}X>{\centering\arraybackslash}X>{\centering\arraybackslash}X}
    \toprule
    \multicolumn{1}{c|}{\multirow{2}{*}{{Captioning Model}}} & CaptionQA & CLIPImage & CLIPScore & FLEUR & CTTD \\
     & \citeyear{caprl} & \citeyear{VisualFactChecker} & \citeyear{clipscore} & \citeyear{fleur} & (Ours) \\
    \midrule
    BLIP~\cite{li2022blip}  & 59.56  & 77.79  & 18.89  & 48.39  & 48.59  \\
    LLaVa-V1.6-7B~\cite{liu2024improved} & 58.22  & 91.53  & 26.28  & 48.23  & 59.77  \\
    InternVL3.5-8b~\cite{internvl3_5} & \underline{67.09}  & 92.73  & 26.49  & 49.38  & 61.53  \\
    Qwen2.5-VL-3B~\cite{qwen2_5_vl} & 64.74  & 92.11  & 26.29  & 48.45  & 60.96  \\
    Qwen2.5-VL-7B~\cite{qwen2_5_vl} & 66.01  & 92.81  & 26.56  & 49.32  & 61.43  \\
    Qwen3-VL-2B~\cite{qwen3_vl} & 66.89  & \underline{92.91}  & \textbf{26.84 } & \underline{49.42}  & \underline{61.79 } \\
    Qwen3-VL-32B~\cite{qwen3_vl} & \textbf{70.12 } & \textbf{93.46 } & \underline{26.81}  & \textbf{49.72 } & \textbf{62.55 } \\
    \bottomrule
\end{tabularx}%
  \caption{Scores from different caption evaluations across captioning models. Bold numbers indicate the best performance, and underlined numbers indicate the runner-up. The methods disagree on several model pairs.
}
  \label{tab:supp_metrics}%
\end{table*}

\subsection{Result Analysis}

\noindent\textbf{Q1.} \textbf{\textit{What patterns does the framework recover?}}

\noindent\textbf{A1.}
We report three internal consistency patterns observed in the experiments; none substitutes for direct correlation with human judgments of caption quality.
First, traditional captioning models exhibit weaker performance, whereas the evaluated VLMs and LVLMs show progressively higher scores across the task suite. As shown in Table~\ref{tab:vqa}, ShowAndTell averages 20.78 and BLIP averages 28.62. LVLMs benefit from richer training data and larger model capacity, which can support captions that preserve more of the information queried by the downstream tasks: even LLaVa-V1.6 7B reaches 38.83, an 86\% increase over ShowAndTell, while Qwen3-VL 32B is highest among the evaluated models at 43.73. This ordering is an internal sanity check, not construct validation.

Second, within each evaluated model family, score tends to increase with scale, suggesting that larger models capture more of the visual information needed to produce detailed captions. 
For Qwen2.5-VL, scores rise from 40.43 to 42.65 and 42.97 for 3B, 7B, and 72B, respectively. The InternVL3.5, Qwen3-VL, and LLaVa-V1.6 families exhibit the same within-family trend. This observation alone does not establish that publication date or scaling~\cite{kaplan2020scaling} determines caption quality.

Lastly, because the difficulty of different vision-language tasks varies, the gap between model performance and theoretical upper bounds also varies. 
Third, task difficulty and information requirements lead to different gaps from the original-image upper reference. The best-performing models for VQA, STEM, OCR, and Grounding are Qwen3-VL 32B, InternVL3.5 38B, Qwen3-VL 2B, and Qwen3-VL 32B, with average performances of 61.03, 30.13, 41.46, and 23.65, respectively. The corresponding best-model gaps are 4.74 for VQA and 3.27 for STEM, versus 19.98 for OCR and 13.30 for Grounding. VQA and STEM focus more on semantic understanding, for which captions can retain a substantial portion of the relevant content. In contrast, OCR and Grounding require low-level text or localization information that caption compression can omit. These differences are therefore consistent with captions preserving semantics while losing fine details, but the OCR and Grounding gaps also include generator, judge, and stochastic-layout error.

\noindent\textbf{Q2.} \textbf{\textit{Does CTTD approximate the full task suite?} }

\noindent\textbf{A2.} 
Figure~\ref{fig:CTTD_overall}(a) shows that CTTD reproduces the broad performance trends of the full vision-language task suite in Table~\ref{tab:vqa}, supporting its use as a computational surrogate within this setup but not establishing agreement with human judgments. CTTD is approximately ten times faster than exhaustive task evaluation. Qwen3-VL 32B achieves the highest average score of 64.3, while LLaVa-V1.6 7B, the lowest-scoring evaluated LVLM, reaches 61.2 versus ShowAndTell's 50.5, a 21\% improvement. Thus, the separation between traditional captioners and the evaluated LVLMs is also visible in this lower-cost evaluation. Within Qwen2.5-VL, performance rises from 62.4 for 3B to 63.5 for 72B.

In terms of question categories, Figure~\ref{fig:CTTD_overall}(b) shows that Qwen3-VL 32B achieves the best performance across most categories, including 60.52 in Chart and 56.36 in Comparison. 
The best models in Commonsense, Explanation, and Temporal are Qwen3-VL 2B, Qwen2.5-VL 72B, and Qwen2.5-VL 3B, with scores of 85.23, 88.98, and 63.39, respectively. As expected in \textbf{Q1.}, categories requiring image details show a large gap to the upper bound: Grounding reaches 67.54, 15.7\% above the best model. Traditional captioners remain competitive in simpler categories such as Commonsense and Explanation.

\noindent\textbf{Q3. \textit{How does our caption semantic evaluation differ from other semantic methods?}
} 


\noindent\textbf{A3.}
Our task-driven framework uses question categories spanning multiple vision-language capabilities, making the evaluation more comprehensive than a single embedding similarity. Like other automatic metrics, however, it inherits errors from its core components: embedding metrics depend on embedders, while reconstruction metrics depend on generators, judges, and question selection.
The reconstructed image is not compared with the source at the pixel level; instead, it mediates task probes and gives the judge visual evidence unavailable to caption-only QA methods such as CaptionQA~\cite{caprl}. This broader semantic coverage is the principal design difference, but it does not remove correlated component errors. 

Our framework also avoids dependence on reference captions and their annotator-specific content selection. Table~\ref{tab:supp_metrics} illustrates metric disagreement: CaptionQA and FLEUR rank LLaVa-V1.6 7B above BLIP, whereas CLIPScore ranks Qwen2.5-VL 3B above 7B. Caption granularity is one plausible source of the first disagreement because LLaVa-V1.6 7B produces more detailed captions while BLIP produces briefer descriptions. The second disagreement departs from the within-family scaling trend observed in Table~\ref{tab:vqa}. These cases are diagnostic rather than proof that any metric is correct; resolving them requires correlation and error analysis against human judgments.
See the supplement for additional analyses.

\section{Conclusion}
\label{sec:conclusion}

This work studies semantic reconstruction as a reference-free signal for image-caption evaluation. We operationalise the principle by testing a caption-conditioned reconstruction across multiple vision-language tasks, yielding a task-based diagnostic rather than a universal definition of caption quality.
We also introduce the Captioning Turing Test Dataset (CTTD), a lower-cost surrogate for the full task suite. Experiments show similar broad model trends under the two evaluations, but these trends are internal consistency evidence: ordering systems by scale or release date does not establish agreement with human judgments. Direct correlation with detailed human caption assessments remains unmeasured and is therefore the central validation still required. The practical score in Eq.~\eqref{eq:vl_score} should likewise be distinguished from the equivalence criterion in Eq.~\eqref{eq:vl_equiv}. Because it uses reconstructed-image accuracy directly, a simplified or standardized image can occasionally be easier for the judge even when its caption omits details. Grounding IoU additionally conflates captioned spatial relations with stochastic shifts or mirroring introduced by the generator, while OCR and STEM scores depend on whether both the generator and judge can reproduce and interpret text or diagrams. CTTD reduces computation but does not remove these component effects: generator--judge correlations and embedding-based QA selection can favor questions that the pipeline handles reliably. Accordingly, the framework complements rather than replaces human evaluation and reference-based metrics, and its scores should be interpreted under fixed pipeline components.
Future work should measure system- and caption-level human correlation on contemporary detailed-caption benchmarks; vary generators and judges; report random success and failure cases; test text-only QA controls; and quantify runtime, question-selection bias, stochastic layouts, and simplification bias before model optimization.

\bibliographystyle{unsrtnat}
\bibliography{main}

\clearpage
\appendix
\section{Supplementary Material}

The following supplementary materials provide additional details to complement the main content. \textbf{More details for the method:} We define the evaluation metrics for various vision-language tasks and formalize some aspects of image caption evaluation. \textbf{More details for implementation:} We offer more detailed implementation information, including dataset descriptions, prompts, and other relevant details. \textbf{More details for experiments:} We supplement the results in the CTTD table, including additional ablation experiments and case analyses.

\subsection*{Method}

\subsubsection*{Vision-Language Task Metrics}

\noindent \textbf{Accuracy}.
For the three vision-language tasks: General Visual Question Answering (VQA), Science, Technology, Engineering, and Mathematics (STEM), and Optical Character Recognition (OCR), we use accuracy as the evaluation metric. Accuracy is defined as the ratio of correctly predicted instances to the total number of instances.

\noindent \textbf{Intersection over Union~(IoU)}~\cite{yu2016unitbox}.
For the visual grounding task, we use Intersection over Union (IoU)~\cite{yu2016unitbox} as the evaluation metric. Its definition is as follows:

\begin{equation*}
    \text{IoU}(\hat{B},B)=\frac{| \hat{B} \cap B |}{| \hat{B} \cup B |}
\end{equation*}
where \( \hat{B} \) and \( B \) are the predicted and ground truth bounding boxes, respectively. \( | \hat{B} \cap B | \) is the area of intersection between the predicted and ground truth bounding boxes.  \( | \hat{B} \cup B | \) is the area of the union of the predicted and ground truth bounding boxes.

\subsubsection*{Image Caption Evaluation}
\noindent \textbf{CaptionQA}~\cite{qiao2024prism, caprl} uses the accuracy of answering visual questions based on the caption to evaluate the accuracy and comprehensiveness of the caption. The basic assumption is that if a caption is sufficiently accurate and complete, it should support answering several questions about the image, with the caption providing correct answers. This is formalized as follows:
\begin{equation*}
    S_{\text{CaptionQA}} = \frac1{k} \sum_{i=1}^{k} \mathbb I[a_{i}=\mathcal{J}(C,q_i)]
\end{equation*}
where $\{(q_i, a_i)\}$ represents the visual question-answer pairs, and $k$ denotes the number of visual questions, and $\mathcal{J}$ is the judge model that answers the question based on the caption $C$. $\mathbb{I}(\cdot)$ is an indicator function that takes the value 1 when the condition is true, and zero otherwise.

\noindent \textbf{FLEUR}~\cite{fleur} uses the LVLM to compare the matching degree between the image and caption directly, and outputs a continuous score $S_{\text{FLEUR}} \in [0,1]$, along with a natural language explanation indicating the basis for the score.

\noindent \textbf{CLIP-Image-Score}~\cite{VisualFactChecker} reconstructs the caption $C$ using an image reconstruction model (such as Stable Diffusion~\cite{rombach2022high}), obtaining $\hat{I}$, and then calculates the visual similarity between the original image $I$ and $\hat{I}$ in the CLIP embedding space~\cite{clip}. The formula is as follows:
\begin{equation*}
    S_{\text{CLIPImage}} = \cos\bigl(\mathcal{E}(I),\mathcal{E}(\hat{I}))
\end{equation*}
where $\mathcal{E}$ represents the CLIP embedding model, and $\hat{I}$ is the image reconstructed from the caption $C$ using an image reconstruction model.

\noindent \textbf{CLIPScore}~\cite{clipscore} directly computes the cosine similarity between the image embedding and the caption embedding in the CLIP space, without requiring human reference. The formula is as follows:
\begin{equation*}
    S_{\text{CLIPScore}} = \max(0,\cos(\mathcal{E}(I),\mathcal{E}(C)))
\end{equation*}

\begin{figure*}[!t]
    \centering
    \includegraphics[width=\linewidth]{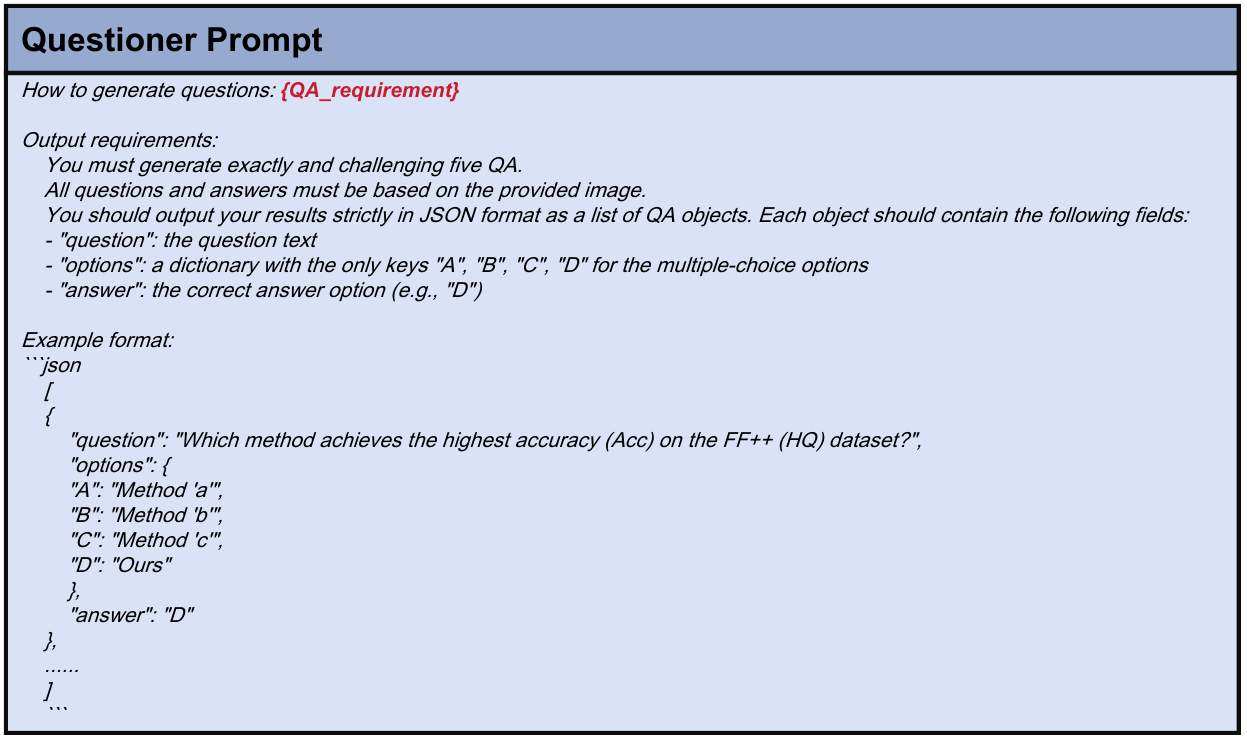}
    \caption{ Prompts for generating questions of each category. The red portions indicate the parts that will be replaced with specific question prompts during implementation, as shown in Figure~\ref{fig:supp_QA_category}.}
    \label{fig:supp_questioner}
\end{figure*}
\subsection*{Implementations}

\subsubsection*{Datasets}
We will provide a detailed introduction to the datasets mentioned for various vision-language tasks in the main paper.

\noindent\textbf{VQA} includes the following four datasets:
\begin{itemize}
    \item MMBench~(MMB.)~\cite{mmbench} features bilingual (i.e., English and Chinese) questions, multiple-choice question formats, and covers a broad range of capabilities, providing a versatile benchmark for testing the performance of vision-language models in various scenarios, from basic question answering to more complex visual reasoning.
    \item MMStar~(MMS.)~\cite{mmstar} is designed for vision-indispensable scenarios, ensuring that each sample requires visual content for answering. The dataset consists of 1500 samples and is intended to test models’ ability to rely on visual information rather than linguistic clues alone.
    \item HallusionBench~(Hall.)~\cite{hallusionbench} focuses on hallucination and illusion in LVLMs. The dataset includes 951 samples and is designed to assess the model’s ability to accurately represent visual content without generating misleading or fabricated details.
    \item MME benchmark~(MME)~\cite{mme} covers the perceptual and cognitive abilities of LVLMs, with 14 sub-tasks (including existence judgment, counting, color recognition, etc.), providing a comprehensive evaluation of how well these models handle various challenges in visual contexts.
\end{itemize}

\noindent\textbf{STEM} includes the following three datasets:
\begin{itemize}
    \item MathVista~(Vista.)~\cite{mathvista} consists of a set of multimodal math tasks, including 6141 problems that combine 28 public datasets and newly created datasets. It is designed to challenge models in solving complex mathematical problems that integrate both textual and visual elements.
    \item MathVerse~(Verse.)~\cite{mathverse} focuses on visual math problems (e.g., geometric shapes, function graphs), with 2612 math problems accompanied by illustrations. The dataset is designed to evaluate a model’s ability to reason about visualized mathematical content and solve problems that require both visual and symbolic understanding.
    \item MathVision~(Vision.)~\cite{MathVision} consists of real-world math competition problems, with 3040 math questions covering 16 mathematical disciplines and five difficulty levels. This dataset tests models’ ability to tackle real-world, competitive mathematical problems ranging from basic arithmetic to more advanced concepts.
\end{itemize}

\noindent\textbf{OCR} includes the following two datasets:
\begin{itemize}
    \item OCRBench~(OCR.)~\cite{ocrbench} includes text recognition and key information extraction, covering 29 sub-datasets. It serves as a comprehensive resource for testing OCR models in diverse real-world scenarios, from document scanning to handwritten text recognition and extraction.
    \item OCRVQA~\cite{ocrvqa} involves answering questions by reading text in images; it consists of approximately 207,572 images, with 5\% sampled for evaluation. This dataset combines text extraction and question answering, allowing models to demonstrate their ability to read and comprehend textual content in a variety of visual contexts.
\end{itemize}
\begin{figure*}[!ht]
    \centering
    \includegraphics[width=\linewidth]{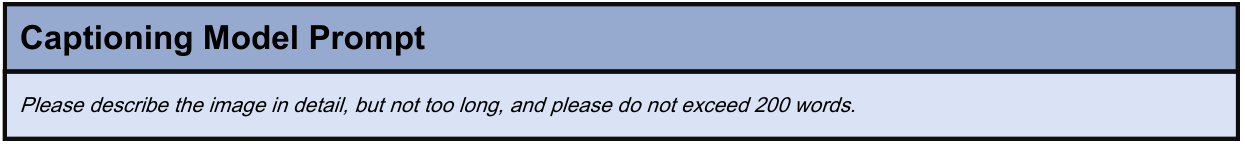}
    \caption{Prompt for the captioning model to generate captions for an image. We adjust the caption length by restricting the caption generation length in the prompt.}
    \label{fig:supp_captioner}
\end{figure*}

\begin{figure*}[!ht]
    \centering
    \includegraphics[width=\linewidth]{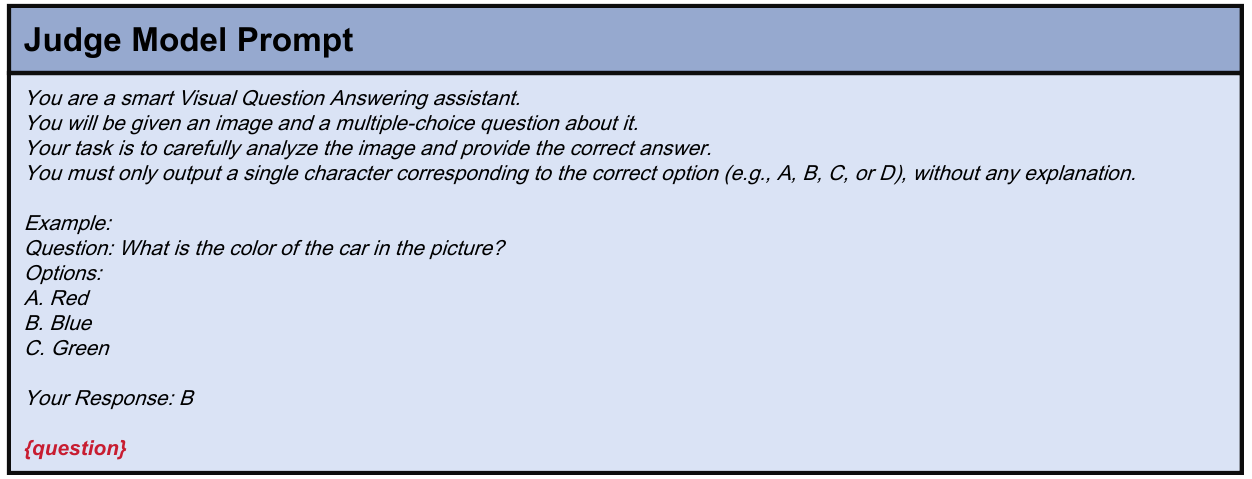}
    \caption{Prompt for the judge model to answer visual questions by reconstructing the image. The red portions will be replaced with specific visual questions during implementation.}
    \label{fig:supp_juder}
\end{figure*}
\noindent\textbf{Visual Grounding}. The representative dataset is RefCOCO~(COCO.)~\cite{refcoco}, which includes each sample with a grounding box and several object descriptions. This dataset is used to evaluate a model’s ability to link specific image regions to natural language descriptions, testing its visual grounding.

\begin{figure*}[!t]
    \centering
    \includegraphics[width=\linewidth]{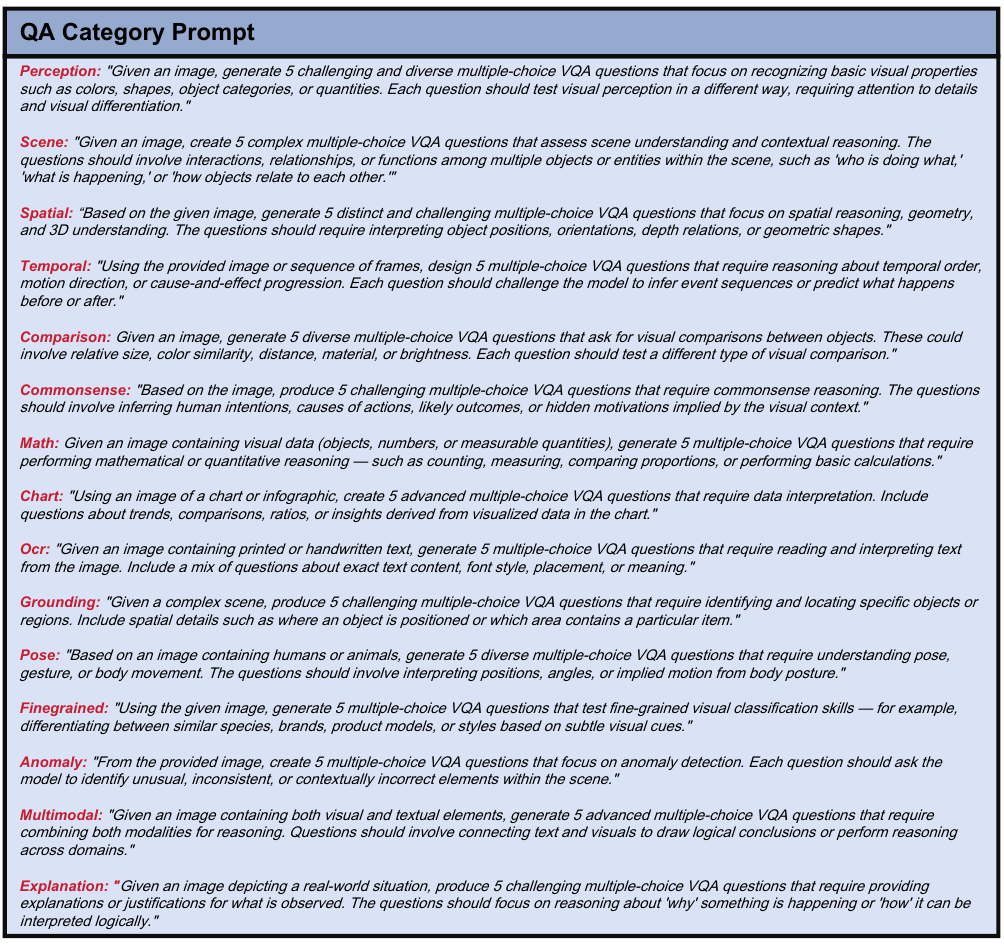}
    \caption{Prompt for generating 15 categories of questions, including Perception, Scene, Spatial, Temporal, Comparison, Commonsense, Math, Chart, OCR, Grounding, Pose, Fine-grained, Anomaly, Multimodal, and Explanation.}
    \label{fig:supp_QA_category}
\end{figure*}

\subsubsection*{QA Category}
Below is a detailed introduction to the 15 question categories in the CTTD. In Figure~\ref{fig:supp_QA_category} and~\ref{fig:supp_questioner}, we also present the prompts used to generate questions for each category.
\begin{itemize}
    \item \textbf{Perception}: Basic visual perception and object recognition questions (identifying quantities, colors, shapes, and objects)
    \item \textbf{Scene}: Scene understanding and contextual reasoning questions (recognizing visual relationships among multiple objects or entities)
    \item \textbf{Spatial}: Spatial and geometric reasoning questions (understanding position, rotation, orientation, or 3D structure; inferring relative depth, volume, or spatial arrangement from images)
    \item \textbf{Temporal}: Temporal and sequential reasoning questions (inferring the order of events, motion, or time-based changes)
    \item \textbf{Comparison}: Visual comparison and matching questions (comparing size, distance, materials, or other visual attributes)
    \item \textbf{Commonsense}: Visual commonsense reasoning questions (inferring plausible causes, effects, or intentions from visual cues)
    \item \textbf{Math}: Mathematical and quantitative reasoning questions (solving arithmetic, algebraic, or data-based problems derived from visual information)
    \item \textbf{Chart}: Graph and chart interpretation questions (analyzing bar charts, pie charts, infographics, or statistical visuals)
    \item \textbf{OCR}: Optical character recognition and text localization questions (detecting and reading text regions or handwritten elements in images)
    \item \textbf{Grounding}: Object detection and localization questions (identifying and locating specific objects or bounding boxes in a scene)
    \item \textbf{Pose}: Pose estimation and motion reasoning questions (identifying human or animal keypoints, gestures, or body postures)
    \item \textbf{Finegrained}: Fine-grained visual categorization questions (distinguishing detailed subcategories such as species, brands, or object models)
    \item \textbf{Anomaly}: Visual anomaly and outlier detection questions (detecting abnormal, inconsistent, or unexpected visual elements within a scene)
    \item \textbf{Multimodal}: Multimodal reasoning and fusion questions (integrating visual, textual, and symbolic information to answer complex or cross-domain queries)
    \item \textbf{Explanation}: Commonsense-based visual explanation questions (providing reasoning or justification for observed visual situations or events)
\end{itemize}

\subsubsection*{Prompts}
As shown in Figure~\ref{fig:supp_captioner} and \ref{fig:supp_juder}, these are the prompts for the captioning model and the judge model, respectively.

\begin{table*}[!t]
  \centering
    \renewcommand{\arraystretch}{1.2}
    \scalebox{0.83}{
\begin{tabularx}{1.2\textwidth}{l|>{\centering\arraybackslash}X|>{\centering\arraybackslash}X>{\centering\arraybackslash}X>{\centering\arraybackslash}X>{\centering\arraybackslash}X>{\centering\arraybackslash}X>{\centering\arraybackslash}X>{\centering\arraybackslash}X>{\centering\arraybackslash}X>{\centering\arraybackslash}X>{\centering\arraybackslash}X>{\centering\arraybackslash}X>{\centering\arraybackslash}X}
    \toprule
    Captioning Model & \textbf{Overall} & Anomaly & Chart & \makecell{Common-\\sense} & \makecell{Compar-\\ison} & Grounding & \makecell{Explana-\\tion} & \makecell{Fine-\\grained} \\
    \midrule
    Upperbound & 68.99  & 69.88  & 63.70  & 86.09  & 61.50  & 67.54  & 89.80  & 66.26  \\
    \midrule
    ShowAndTell~\cite{vinyals2015show} & 50.47  & 60.70  & 42.94  & 79.47  & 42.15  & 40.49  & 83.99  & 39.62  \\
    BLIP~\cite{li2022blip}  & 57.89  & 64.76  & 53.52  & 82.58  & 49.61  & 49.05  & 86.46  & 48.20  \\
    BLIP2~\cite{li2023blip} & 58.61  & 64.91  & 54.68  & 82.64  & 50.32  & 51.62  & 86.63  & 49.62  \\
    LLaVa-V1.6-7B~\cite{liu2024improved} & 61.23  & 66.69  & 57.01  & 83.68  & 53.39  & 53.78  & 87.82  & 52.86  \\
    LLaVa-V1.6-34B~\cite{liu2024improved} & 61.97  & \textbf{67.55 } & 58.10  & 83.74  & 54.98  & 54.52  & 88.37  & 52.94  \\
    Llama-3.2-11B~\cite{llama3_2_11b_vision} & 62.87  & \underline{67.53}  & 59.22  & 84.25  & 54.62  & 55.82  & 88.21  & 55.60  \\
    Qwen2.5-VL-3B~\cite{qwen2_5_vl} & 62.44  & 66.93  & 58.89  & 83.82  & 54.49  & 55.20  & 88.15  & 54.98  \\
    Qwen2.5-VL-7B~\cite{qwen2_5_vl} & 62.99  & 67.12  & 59.21  & 84.34  & 54.64  & 56.02  & 88.89  & 55.20  \\
    Qwen2.5-VL-72B~\cite{qwen2_5_vl} & 63.53  & 67.28  & 59.68  & \underline{84.96}  & 55.38  & 56.70  & \textbf{88.98 } & 55.81  \\
    InternVL3.5-8b~\cite{internvl3_5} & 63.27  & 67.06  & 59.71  & 84.95  & 54.92  & 56.50  & 88.53  & 55.05  \\
    InternVL3.5-38b~\cite{internvl3_5} & 63.44  & 66.86  & \underline{59.79}  & 84.51  & \underline{55.61}  & 56.51  & 88.91  & 56.54  \\
    Qwen3-VL-2B~\cite{qwen3_vl} & \underline{63.62}  & 67.25  & 59.44  & \textbf{85.23 } & 54.60  & \underline{57.37}  & 88.80  & 56.21  \\
    Qwen3-VL-32B~\cite{qwen3_vl} & \textbf{64.29 } & 67.46  & \textbf{60.52 } & 84.92  & \textbf{56.36 } & \textbf{58.33 } & 88.80  & \textbf{58.04 } \\
    \bottomrule
\end{tabularx}%
    }
  \caption{Performance of different captioning models in CTTD. Bold numbers indicate the best performance, and underlined numbers indicate the runner-up. ``Upperbound'' refers to the performance achieved by completing the task using the original image.
}
  \label{tab:supp_CTTD_Full_1}%
\end{table*}

\begin{table*}[!t]
  \centering
    \renewcommand{\arraystretch}{1.2}
    \scalebox{0.83}{
\begin{tabularx}{1.2\textwidth}{l|>{\centering\arraybackslash}X>{\centering\arraybackslash}X>{\centering\arraybackslash}X>{\centering\arraybackslash}X>{\centering\arraybackslash}X>{\centering\arraybackslash}X>{\centering\arraybackslash}X>{\centering\arraybackslash}X>{\centering\arraybackslash}X>{\centering\arraybackslash}X>{\centering\arraybackslash}X>{\centering\arraybackslash}X>{\centering\arraybackslash}X}
    \toprule
    Captioning Model & Math  & \makecell{Multi-\\modal} & OCR   & Perception & Pose  & Scene & Spatial & Temporal \\
    \midrule
    Upperbound & 44.91  & 69.75  & 75.38  & 66.58  & 70.87  & 70.53  & 56.63  & 68.03  \\
    \midrule
    ShowAndTell~\cite{vinyals2015show} & 26.36  & 55.96  & 37.59  & 32.96  & 46.47  & 55.13  & 39.41  & 53.88  \\
    BLIP~\cite{li2022blip}  & 40.30  & 62.14  & 49.54  & 42.58  & 56.36  & 62.28  & 46.61  & 59.46  \\
    BLIP2~\cite{li2023blip} & 40.81  & 62.78  & 49.49  & 43.51  & 57.92  & 62.83  & 47.35  & 59.48  \\
    LLaVa-V1.6-7B~\cite{liu2024improved} & 42.27  & 64.84  & 56.76  & 49.33  & 60.15  & 65.54  & 49.45  & 61.54  \\
    LLaVa-V1.6-34B~\cite{liu2024improved} & 43.65  & 65.15  & 58.02  & 50.15  & 60.54  & 66.41  & 50.32  & 61.97  \\
    Llama-3.2-11B~\cite{llama3_2_11b_vision} & 43.56  & 65.55  & 60.89  & 53.14  & 62.48  & 66.58  & 50.66  & 62.81  \\
    Qwen2.5-VL-3B~\cite{qwen2_5_vl} & 43.95  & 65.64  & 58.60  & 51.82  & 61.26  & 66.25  & 50.66  & \textbf{63.39 } \\
    Qwen2.5-VL-7B~\cite{qwen2_5_vl} & 42.57  & 65.96  & 59.79  & 53.79  & 62.64  & 66.88  & 51.22  & 63.23  \\
    Qwen2.5-VL-72B~\cite{qwen2_5_vl} & \underline{44.40}  & 66.81  & 62.66  & 53.88  & 63.19  & 66.99  & \underline{51.68}  & 63.28  \\
    InternVL3.5-8b~\cite{internvl3_5} & 43.62  & 66.60  & 63.04  & 53.21  & 62.75  & 67.03  & 51.30  & 63.32  \\
    InternVL3.5-38b~\cite{internvl3_5} & 43.99  & \underline{66.84}  & 62.63  & 52.87  & 63.67  & 67.04  & 51.54  & 63.00  \\
    Qwen3-VL-2B~\cite{qwen3_vl} & 44.14  & 66.55  & \underline{63.55}  & \underline{54.43 } & \underline{63.74}  & \underline{67.14 } & 51.47  & \underline{63.32}  \\
    Qwen3-VL-32B~\cite{qwen3_vl} & \textbf{45.17 } & \textbf{66.97 } & \textbf{65.23 } & \textbf{55.60 } & \textbf{64.82 } & \textbf{67.24 } & \textbf{52.01 } & 63.24  \\
    \bottomrule
\end{tabularx}%
    }
  \caption{Performance of different captioning models in CTTD. Bold numbers indicate the best performance, and underlined numbers indicate the runner-up. ``Upperbound'' refers to the performance achieved by completing the task using the original image.
}
  \label{tab:supp_CTTD_Full_2}%
\end{table*}

\subsection*{Experiments}
\subsubsection*{More Results}

\noindent\textbf{Full Table of CTTD}. As shown in Tables~\ref{tab:supp_CTTD_Full_1} and \ref{tab:supp_CTTD_Full_2}, we provide the performance of different captioning models in CTTD. The performance of models evaluated on CTTD aligns with intuition and common sense. The best-performing model, Qwen3-VL 32B, achieves an average score of 64.3. LVLMs show a clear advantage over traditional captioning models, with the worst-performing LLaVa-V1.6 7B scoring 61.2, which is a 21\% improvement over ShowAndTell. As model size increases (e.g., Qwen2.5-VL series from 3B to 72B, with performance increasing from 62.4 to 63.5), performance improves gradually. In the fine-grained categories, Qwen3-VL 32B achieves the best performance across most dimensions, including Chart (60.52) and Comparison (56.36). At the same time, the SOTA models for Commonsense, Explanation, and Temporal are Qwen3-VL 2B, Qwen2.5-VL-72B, and Qwen2.5-VL-3B, with performance scores of 85.23, 88.98, and 63.39, respectively. For Grounding questions that require image details, the models still show a significant gap relative to the upper bound. Traditional captioning models perform well in some simpler categories, such as Commonsense and Explanation.

\begin{figure*}
    \centering
    \includegraphics[width=\linewidth]{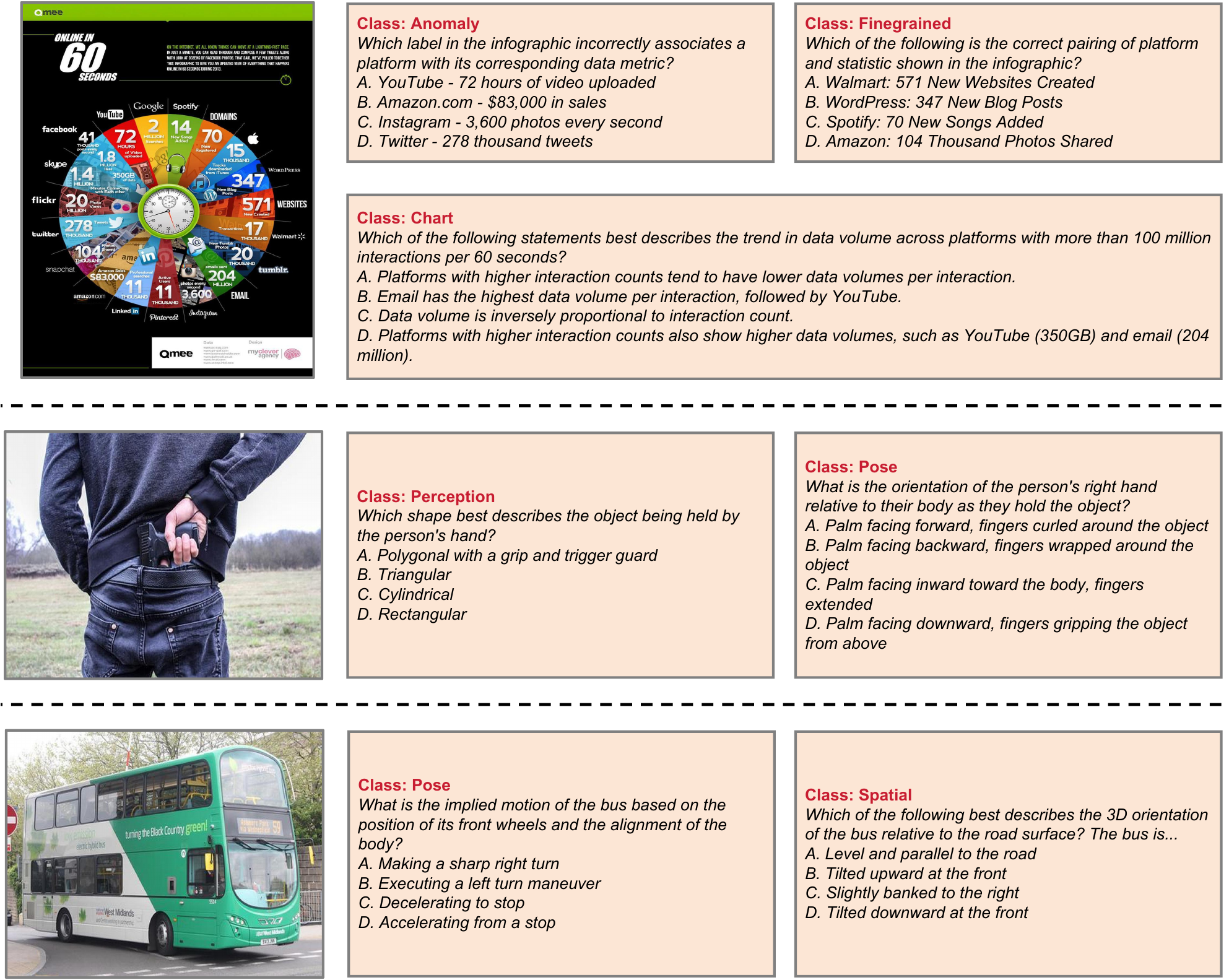}
    \caption{Some case examples in CTTD. The red portions indicate the category of the respective question.}
    \label{fig:supp_case}
\end{figure*}

 \noindent \textbf{Case Study}. Each sample in CTTD contains a rich set of QA pairs, with an average of 15 per sample and 10 distinct question types. Some cases are shown in Figure~\ref{fig:supp_case}. From these QA pairs, it is evident that each QA is quite accurate, highly relevant to the image, and has minimal content redundancy between the questions and answers.

\begin{figure}
    \centering
    \includegraphics[width=1\linewidth]{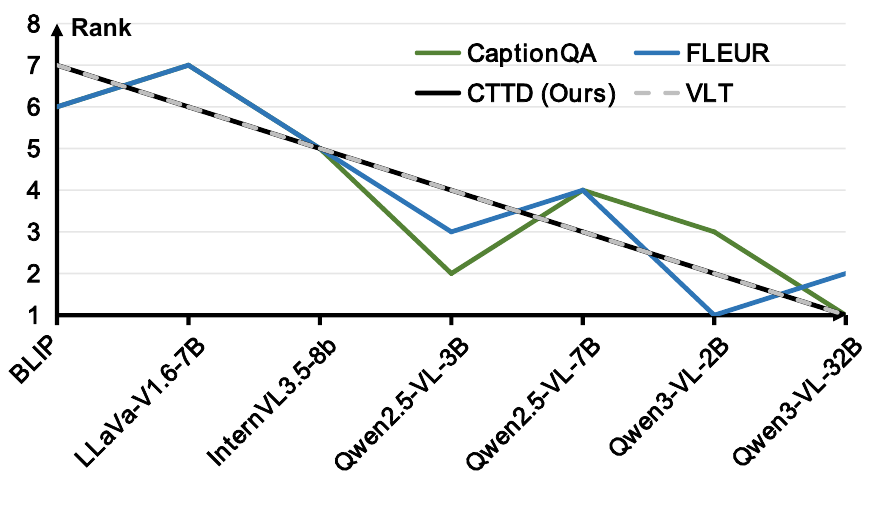}
    \caption{
    Comparison of ranking performance across different evaluation methods. The figure shows the ranks of various models according to CaptionQA~\cite{caprl}, FLEUR~\cite{fleur}, CTDD (ours), and VLT~(vision-language tasks) evaluation.
    }
    \label{fig:metrics_rank}
\end{figure}

\noindent \textbf{Caption Evaluation Comparison}. Furthermore, as shown in Figure~\ref{fig:metrics_rank}, CTTD demonstrates stronger consistency with vision-language task evaluations. Interestingly, the CaptionQA and FLEUR methods ranked LVLM LLaVa-V1.6-7B below VLM BLIP, contradicting the models' actual performance on captioning tasks.


\subsubsection*{More Discussion}

\noindent \textbf{CTTD Scalability}. The CTTD is designed to be scalable, providing a computationally more efficient alternative to exhaustive testing. The framework performs well across a range of vision-language tasks and delivers consistent results across multiple datasets. Nevertheless, the core architecture of CTTD, including its compact set of question–answer pairs and automatic question categorization, enables adaptation to larger datasets or more complex models.

\noindent \textbf{Performance Across Tasks}. Tasks such as OCR and Grounding, which require pixel-level precision, expose a substantial gap between current state-of-the-art models and the performance upper bound. This gap arises from the inevitable loss of fine-grained details during the compression of visual information into textual descriptions. To improve image captions for these tasks, models may need to incorporate more specialized information, such as higher-resolution image descriptions or more fine-grained image encoding mechanisms.

\noindent \textbf{Generalization of our Framework}. The framework proposed in this paper primarily evaluates image captions using downstream tasks that span a broad spectrum of vision-language capabilities. Regarding its generalization ability to emerging tasks, especially those requiring subjective interpretation or creativity, the framework may require further refinement. Future work could explore how to extend CTTD to such subjective tasks, for instance, by introducing human–AI interactive evaluation or more fine-grained semantic consistency checks.

\noindent \textbf{Bias in Judge Models}. Any inherent biases in the judge model—whether linguistic, visual, or otherwise—must be carefully considered. Our framework evaluates a broad range of vision-language tasks, and the CTTD dataset features diverse data sources and question categories, helping mitigate judge-model hacking. As a result, the relative performance differences observed under our framework provide a reliable reflection of captioning models’ current capabilities.

\begin{table}[t]
  \centering
    \renewcommand{\arraystretch}{1.2}
    \resizebox{\columnwidth}{!}{%
\begin{tabular}{l|ccccc}
    \toprule
    \multicolumn{1}{c|}{\multirow{2}{*}{{Generator}}} & BLIP  & LLaVa & InternVL & Qwen2.5 & Qwen3 \\
     & \cite{li2022blip}  & \cite{liu2024improved} & \cite{internvl3_5} & \cite{qwen2_5_vl} & \cite{qwen3_vl}\\
    \midrule
    Qwen~\cite{wu2025qwenimagetechnicalreport}  & 47.19  & 57.68  & 58.64  & 58.81  & 59.44  \\
    SD3.5~\cite{stabilityai2024stable} & 49.85  & 58.90  & 59.30  & 60.01  & 60.29  \\
    \bottomrule
\end{tabular}%
    }
\caption{
Ablation study of different generators on the Captioning Turing Test Dataset. The table compares the performance of two generators, Qwen (Qwen-Image) and SD3.5 (StableDiffusion-3.5-Large-Turbo), across various models: BLIP, LLaVa-V1.6-7B, InternVL3.5-8B, Qwen2.5-VL-7B, and Qwen3-VL-32B.
}

  \label{tab:ablation_gen}%

\end{table}

\begin{table}[t]
  \centering
    \renewcommand{\arraystretch}{1.2}
    \resizebox{\columnwidth}{!}{%
\begin{tabular}{l|ccccc}
    \toprule
     \multicolumn{1}{c|}{\multirow{2}{*}{{Length}}} & BLIP  & LLaVa & InternVL & Qwen2.5 & Qwen3 \\
         & \cite{li2022blip}  & \cite{liu2024improved} & \cite{internvl3_5} & \cite{qwen2_5_vl} & \cite{qwen3_vl}\\
    \midrule
        100 & 48.41  & 59.97  & 61.43  & 61.35  & 62.28  \\
        200 & 46.62  & 55.44  & 56.10  & 56.38  & 56.74  \\
        300 & 48.64  & 60.77  & 61.65  & 61.79  & 62.83  \\
        400 & 48.13  & 60.02  & 61.45  & 61.75  & 62.75  \\
    \bottomrule
\end{tabular}%
    }
  \caption{Ablation study of different maximum caption lengths on the Captioning Turing Test Dataset. The models tested are the same as those in Table~\ref{tab:ablation_gen}.
  }
  \label{tab:ablation_length}%

\end{table}

\begin{table}[t]
  \centering
    \renewcommand{\arraystretch}{1.2}
    \resizebox{\columnwidth}{!}{%
\begin{tabular}{l|ccccc}
    \toprule
    \multicolumn{1}{c|}{\multirow{2}{*}{{Judger}}} & BLIP  & LLaVa & InternVL & Qwen2.5 & Qwen3 \\
     & \cite{li2022blip}  & \cite{liu2024improved} & \cite{internvl3_5} & \cite{qwen2_5_vl} & \cite{qwen3_vl}\\
    \midrule
    Qwen2.5~\cite{qwen2_5_vl} & 47.42  & 51.49  & 51.64  & 52.14  & 52.36  \\
    Qwen3~\cite{qwen3_vl} & 48.68  & 59.93  & 61.03  & 61.26  & 62.05  \\
    \bottomrule
\end{tabular}%
    }
  \caption{Ablation study of different judgers on the Captioning Turing Test Dataset. Compares the performance of two judgers, Qwen2.5 (Qwen2.5-VL-3B) and Qwen3 (Qwen3-VL-8B). The models tested are the same as those in Table~\ref{tab:ablation_gen}.
  }
  \label{tab:ablation_judger}%
\end{table}

We conducted extensive ablation experiments. As shown in Tables~\ref{tab:ablation_gen}-\ref{tab:ablation_judger}, modifying key components of our framework (i.e., the generation model, caption length, and judge model) did not significantly alter the relative performance differences among models. The classical VLM BLIP performed poorly, LVLMs showed substantial improvements, and Qwen3-VL 32B still exhibited SOTA performance. This indicates that CTTD is robust and not easily affected by component, providing a stable ranking for models with smaller evaluation resources.
\noindent \textbf{Discussion about human correlation.}
We respectfully argue that our framework operates on a distinct evaluation track, distinct from metrics designed to align with human preferences.
\textbf{(a)} Our objective is to establish a unified, objective standard for semantic retention, rather than optimizing for subjective human preference. We measure the reliability of captions as information transmission channels by using the attainment of the theoretical upper bound as our ``gold standard.''
\textbf{(b)} Human evaluation often fails to distinguish between fluency and semantics. By quantifying captions' capacity to retain semantics across vision-language tasks, our framework provides a more objective assessment that {complements human evaluation}.

\end{document}